\pdfoutput=1
\documentclass[11pt]{article}

\usepackage{acl}

\usepackage{times}
\usepackage{latexsym}
\usepackage{colortbl}
\usepackage[T1]{fontenc}
\usepackage[utf8]{inputenc}

\usepackage{microtype}

\usepackage{inconsolata}
\usepackage{booktabs} % Required for nicer horizontal lines
\usepackage{hyperref}
\usepackage{url}
\usepackage{graphicx}
\usepackage{amsmath}
\usepackage{amsthm}
\usepackage{amssymb}
\usepackage{multirow}
\usepackage{setspace}
\usepackage{enumitem}
\usepackage{courier}
\usepackage{bm}
\usepackage{geometry}
\usepackage{algorithm}
\usepackage{algorithmicx}
\usepackage{algpseudocode}
\usepackage[algo2e]{algorithm2e}
\usepackage{makecell}
\usepackage{longtable}
\usepackage[utf8]{inputenc}
\usepackage{array}
\usepackage{listings}

\makeatletter
\renewcommand*{\@fnsymbol}[1]{\ensuremath{\ifcase#1\or \dagger\or *\or \ddagger\or
   \mathsection\or \mathparagraph\or \|\or **\or \dagger\dagger
   \or \ddagger\ddagger \else\@ctrerr\fi}}
\makeatother

\title{Aligning Large Language Models with Human Preferences \\ through Representation Engineering}

\author{Wenhao Liu\thanks{\ \ These authors contributed equally.}, Xiaohua Wang\footnotemark[1], Muling Wu, Tianlong Li, Changze Lv \\
{\bf Zixuan Ling, Jianhao Zhu, Cenyuan Zhang,  Xiaoqing Zheng\thanks{\ \ Corresponding author.}, Xuanjing Huang} \\
  School of Computer Science, Fudan University, Shanghai, China \\
  \texttt{\{whliu22,xiaohuawang22\}@m.fudan.edu.cn} \\
 \texttt{\{zhengxq,xjhuang\}@fudan.edu.cn} \\}

\begin{document}
\maketitle
\begin{abstract}

Aligning large language models (LLMs) with human preferences is crucial for enhancing their utility in terms of helpfulness, truthfulness, safety, harmlessness, and interestingness. 
Existing methods for achieving this alignment often involve employing reinforcement learning from human feedback (RLHF) to fine-tune LLMs based on human labels assessing the relative quality of model responses. 
Nevertheless, RLHF is susceptible to instability during fine-tuning and presents challenges in implementation.
Drawing inspiration from the emerging field of representation engineering (RepE), this study aims to identify relevant representations for high-level human preferences embedded in patterns of activity within an LLM and achieve precise control of model behavior by transforming its representations. 
This novel approach, denoted as Representation Alignment from Human Feedback (RAHF), proves to be effective, computationally efficient, and easy to implement.
Extensive experiments demonstrate the efficacy of RAHF in not only capturing but also manipulating representations to align with a broad spectrum of human preferences or values, rather than being confined to a singular concept or function (e.g. honesty or bias). 
RAHF's versatility in accommodating diverse human preferences shows its potential for advancing LLM performance.
Code is available at \url{https://github.com/LiuAmber/RAHF}.

\end{abstract}

\section{Introduction}

While large language models (LLMs) learn broad-ranging world knowledge and a degree of reasoning proficiency, precise control over their behavior proves challenging due to the unsupervised nature of their pre-training \cite{radford2018improving,radford2019language,brown2020language,bubeck2023sparks,touvron2023LLaMA}.
For each query, instruction-tuned LLMs \cite{wei2021finetuned,chung2022scaling,touvron2023LLaMA} exhibit the capacity to generate multiple responses that are both semantically and syntactically coherent by some sampling techniques. 
% This skill in response generation endows the models with the ability to provide diversity, a critical attribute for chat agents.
% However, many studies \cite{srivastava2022beyond,thoppilan2022lamda,bubeck2023sparks} show that some responses may contain harmful, unethical, socially biased, or even illegal content.
While such ability enables the models to provide diversity that is essential for chat agents, some responses may contain harmful, unethical, socially biased, and negative, even illegal content \cite{srivastava2022beyond,thoppilan2022lamda,bubeck2023sparks,wang2023hallucination}. 
%Although many responses remain within acceptable bounds, preferences may exist among them, leading to a desire to prioritize specific outputs over others.  

\begin{figure}[t]
  \centering
  \setlength{\abovecaptionskip}{1mm}
  \includegraphics[width = 7.7cm]{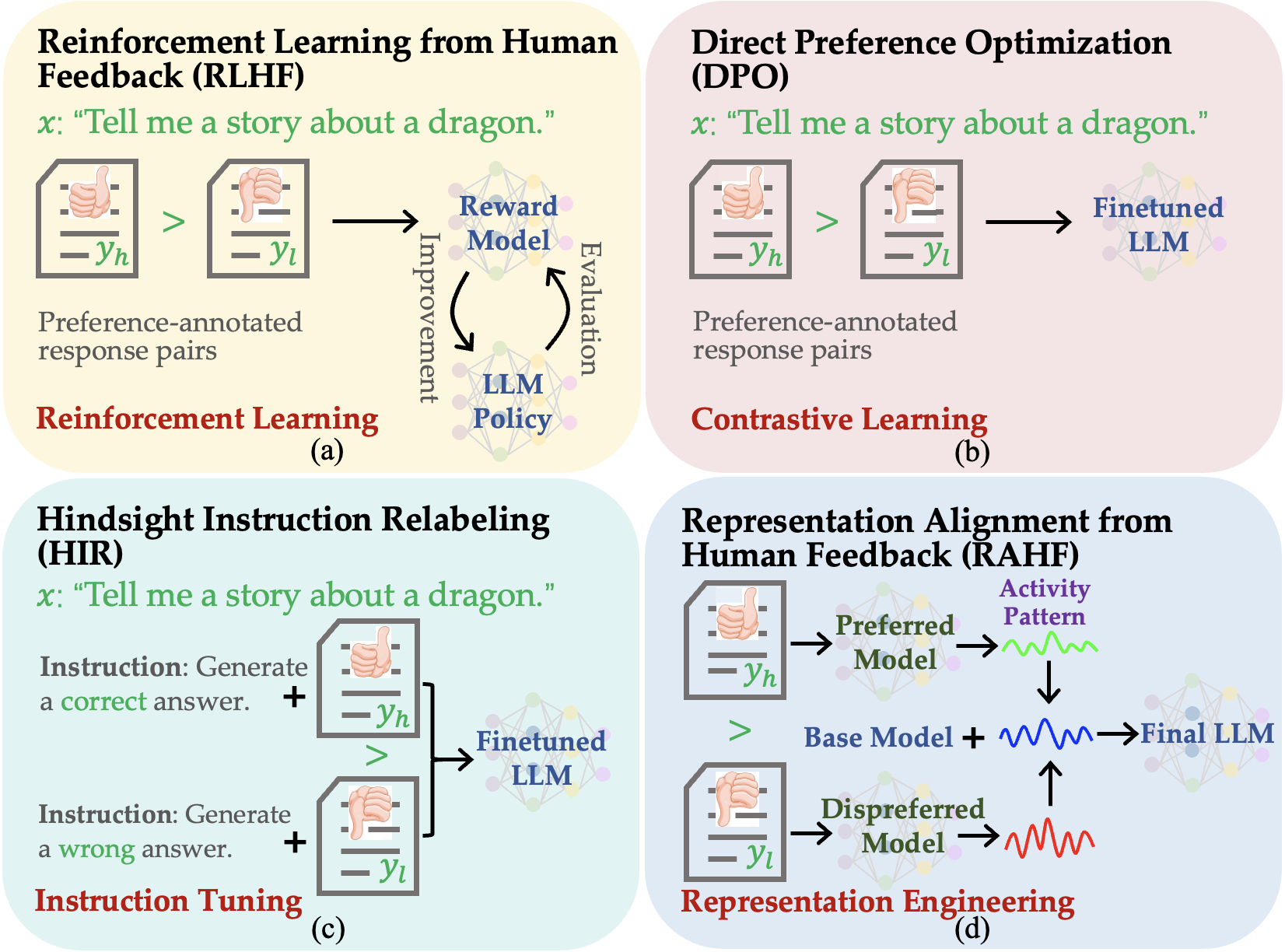}
  \caption{\label{fig:schematic} Illustration of different apporaches.
  (a) Reinforcement learning from human feedback (RLHF); (b) Direct preference optimization (DPO); (c) Hindsight instruction relabeling (HIR); (d) Representation alignment from human feedback (RAHF). }
  \vspace{-6mm}
\end{figure}

Existing methods steer LLMs to align with human preferences often using reinforcement learning (RL), with reinforcement learning from human feedback (RLHF) emerging as the most successful one \cite{ouyang2022training}.
However, the underlying learning algorithms exhibit a considerable degree of complexity, sensitivity to hyperparameters, instability during training, and necessitate additional training in the reward model and value network, leading to substantial computational costs \cite{yuan2023rrhf,rafailov2023direct}.

In addressing the aforementioned challenges posed by RL-based methods, several computationally lightweight alternatives have been proposed to simplify the human preference-matching process.
Two prominent paradigms among these alternatives include contrastive learning \cite{rafailov2023direct,zhao2023slic,yuan2023rrhf} and Hindsight instruction relabeling (HIR) \cite{Zhang-icml23,liu2023languages}.
Contrastive learning-based methods optimize a language model policy by increasing the relative probability of preferred responses over dispreferred ones, while HIR methods transform human feedback into instructions by relabeling the original ones, indicating the relative quality of provided responses.
A common characteristic shared by these two paradigms is their capability to align language models with human preferences through reward-free fine-tuning.

However, the reward-free fine-tuning is vulnerable to the presence of noisy data or incorrect labels in a training set comprising a collection of preference-annotated response pairs \cite{li2023policy,dumoulin2023density}. 
Instances of dull sentences or very brief responses may appear repeatedly in such a training set, potentially introducing bias into the models. 
The exclusion of such instances from the training set renders it impossible for LLMs to glean insights into human preferences expressed in these instances.
% Moreover, the fine-tuning process may inadvertently lead the models to explore more harmful, dangerous, or unethical examples.
In contrast, RL-based methods adopt a different strategy, wherein a reward function is first extracted from a dataset of response rankings, and then this reward function can be applied to train an LLM, effectively mitigating the model's direct exposure to noisy data or incorrect labels within the dataset.

In this study, we aim to seek for a computationally lighter and reward-free algorithm that can effectively harness human preference expressed in datasets meanwhile safeguarding LLMs from the influence of noisy data.
Inspired by the recent advance in representation engineering \cite{zou2023representation}, we initially locate relevant representations and activity patterns associated with high-level human preferences within an LLM, and subsequently gain precise control over its behavior by manipulating its internal representations.
In the neural architecture, network weights determine neural activity, neural activity determines the networks' output, and the networks' output determines the networks' behavior.
Instead of focusing on neurons and their connections, we see aligning LLMs with human feedback as an outcome of representational spaces, implemented by patterns of activity across populations of neurons.
We first identify the differences in model activities between preferred and dispreferred stimuli, and then control its behavior by leveraging the identified differences in representations (see Figure \ref{fig:schematic}). 
We introduce two methods for controlling representations and demonstrate the efficacy of these representation engineering (RepE) approaches in aligning LLMs with a broad spectrum of human preferences through a collection of response pairs.

To validate the effectiveness of our approach in aligning with human preferences, we conducted extensive comparative experiments on the generated results. Our method outperformed RLHF and other RL-free approaches in human evaluations and automated metrics such as general abilities and GPT-4 evaluations.
Notably, the underlying algorithms exhibit simplicity in implementation and straightforwardness in training.
% Gaining precise control of their behavior could greatly reduce the risks posed by LLMs, and existing methods for achieving such steerabilty often finetune

% In particular, we develop improved baselines for controlling representations and demonstrate that these RepE techniques can provide traction on a wide variety of safety-relevant problems.

% in term of human preferences between pairs of responses.

% Representation reading seeks to locate emergent representations for high-level concepts and functions within a network.

% We begin by extracting concepts, including truthfulness, utility, probability, morality, and emotion. 

% Building on the insights gained from representation reading, representation control seeks to modify or control the internal representations of concepts and functions.

\section{Related Work}

Tuning large language models to elicit desired responses and behavior from their extensive knowledge and capabilities is essential in the development of chat agents, such as ChatGPT \cite{brown2020language}, LLaMA \cite{touvron2023LLaMA} and GPT-4 \cite{bubeck2023sparks}, characterized by safety, performance, and controllability.
The enlargement of the size of language models only does not inherently enhance their ability to follow a user’s intent. 
For example, LLMs may still generate outputs that are untruthful, toxic, or simply not helpful to the user.
Existing human preference alignment methods can be broadly classified into three major categories: reinforcement learning \cite{ouyang2022training,ramamurthy2022reinforcement}, contrastive learning \cite{rafailov2023direct,zhao2023slic,yuan2023rrhf}, and Hindsight instruction relabeling \cite{Zhang-icml23,liu2023languages}.
% Averting these unintended behaviors is especially important for language models that are deployed and used in hundreds of applications.

Extensive research has been devoted to the exploration of RL from human feedback through ratings or rankings, spanning tasks from NL-to-SQL conversion \cite{zhong2017seq2sql}, machine translation \cite{kreutzer2018reliability}, task-oriented dialogue systems \cite{su2018discriminative,zhang2019budgeted,takanobu2019guided}, summarization \cite{stiennon2020learning}, story-telling \cite{ziegler2019fine} to instruction-following \cite{ouyang2022training,ramamurthy2022reinforcement}. 
Typically, these methods involve the fitting of a reward model to a dataset of human preferences, followed by the optimization of a LLM policy to generate responses with high reward, using RL algorithms such as REINFORCE \cite{williams1992simple} or proximal policy optimization \cite{schulman2017proximal}.
Despite the attractiveness of leveraging human preferences that are easier to collect than expert demonstrations, training LLMs with RL poses significant practical challenges, which is attributed to the sensitivity of RL to hyperparameters and the inherent instability during training.

The solutions based on Hindsight instruction relabeling \cite{Zhang-icml23,liu2023languages} and contrastive learning \cite{rafailov2023direct,zhao2023slic,yuan2023rrhf} have emerged as computationally efficient alternatives to RL-based methods without explicit reward modeling.
However, these reward-free fine-tuning solutions are susceptible to noisy data or incorrect labels within a training set. 
They exhibit performance lags compared to models tuned with RL counterparts (see Section \ref{sec:experiment}). 
Furthermore, the question of whether LLMs trained with such fine-tuning methods can generalize well to out-of-distribution queries remains unresolved when contrasted with models incorporating an explicit reward model. 
RLHF method \cite{ouyang2022training} offers a potential avenue for improvement by leveraging additional unlabeled examples through labeling LLM generations with the learned reward model.

To enhance transparency and controllability of neural networks, \citet{zou2023representation} introduced representation engineering (RepE) as a methodology, drawing an analogy between understanding deep neural networks through representation tomography and studying brains via neuroimaging techniques. 
Their work demonstrated the efficacy of RepE in addressing diverse safety-related challenges such as truthfulness, honesty, and hallucination. 
This study falls in line with recent research findings and extends its application to aligning LLMs with a wide spectrum of human preferences.
Our study introduces two novel methods to instruct LLMs on human preferences first, and then extract differences in model activities between preferred and dispreferred stimuli. 
These differences in activity patterns serve as a foundation for manipulating the model's behavior, leading to the generation of responses that better align with human preferences.
Due to the lightweight computational advantages of parameter-efficient fine-tuning techniques\cite{houlsby2019parameter, lester2021power, hu2021lora, wu2023parameter,wu2024advancing}, these techniques are utilized to fit the disparity in activity patterns.
In contrast to the approach adopted by \citet{zou2023representation}, which relies on unlabeled or self-generated stimuli limited to singular concepts or functions the meaning of which the models have already ``known'', our methods provide a more comprehensive alignment with diverse human preferences.

% Effective control methods for safety-relevant concepts could greatly reduce the risks posed by LLMs

% Compared to all prior work, our key contribution is to scale human feedback up to deep reinforcement learning and to learn much more complex behaviors.

% that draws on insights from cognitive neuroscience 

\begin{figure*}[!t]
  \centering
  \includegraphics[width=0.80\linewidth]{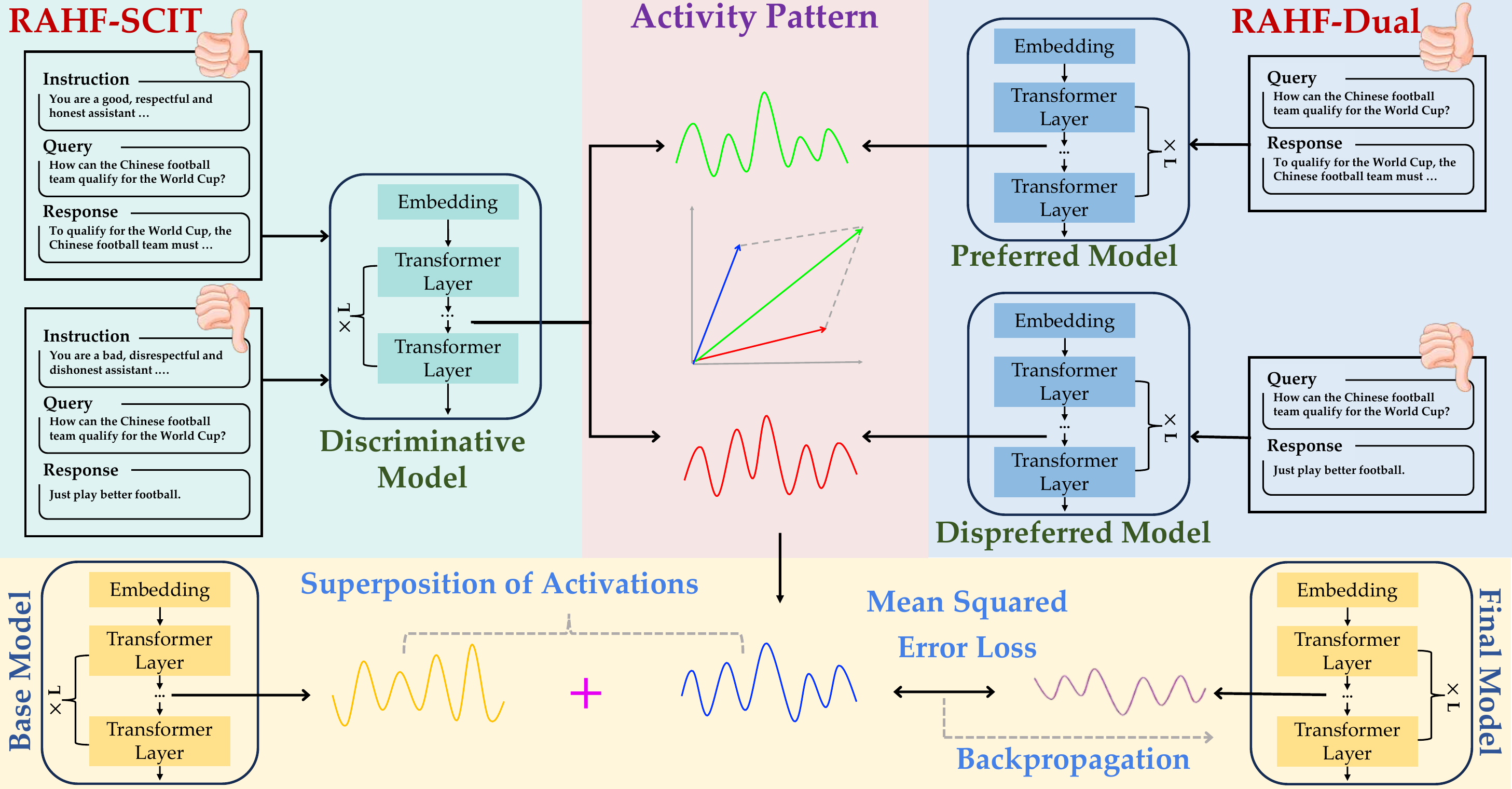}
  \caption{\label{fig:procedure} The procedure of RAHF. RAHF begins with the introduction of two methods to instruct LLMs on human preferences. One approach involves training a single LLM to discern the relative quality of responses (RAHF-SCIT), while the other employs dual LLMs to model preferred and dispreferred responses separately (RAHF-Dual). Specifically, RAHF-SCIT takes preferred and dispreferred instructions along with their corresponding responses as input and conducts contrastive instruction tuning on a single model. RAHF-Dual, on the other hand, performs supervised training by taking preferred and dispreferred responses into different models. Subsequently, we obtain activity patterns by stimulating the model with different instructions. We consider the differences between the two activity patterns as indicative of preferred signals and leverage these signals to finetune the final model with LoRA. }
  \vspace{-3mm}
\end{figure*}

\section{Method}

We begin by instructing LLMs on human preferences with a set of preference-annotated response pairs. 
We introduce two novel methods for instructing LLMs on human preferences and extracting their activity patterns: one involving a single LLM (trained to discern the relative quality of responses) and the other employing dual LLMs (``a good guy and a bad guy'').
Secondly, we collect the activity patterns of LLMs when exposed to stimuli that are preferred or dispreferred. 
The differences in these patterns serve as the foundation for manipulating LLMs, enabling them to generate responses more closely aligned with human values. 
Finally, we construct the final model by training a low-rank adapter\cite{hu2021lora} to fit the disparity in activity patterns.
%, both fine-tuned using parameter-efficient techniques.

\subsection{Instructing LLMs on Human Preferences}
\label{sec:3.1}
To extract activity patterns from the model that align with human preferences, it is crucial for the model to possess a correct understanding and awareness of these preferences. 
The effectiveness of extracting activity patterns from alignment fine-tuned models, such as LLaMA-2-chat, in capturing concepts like truthfulness and honesty has been validated by \citet{zou2023representation}. 
However, for non-aligned models, such as pre-trained large language models or LLMs subjected to simple fine-tuning, explicit indications of human preferences should be provided to elicit and capture activity patterns induced by stimulus preferences. 
This capability enables the accumulation of diverse activities, subsequently utilized to calibrate LLMs based on human preferences.

For instructing LLMs on human preferences, we rely on a dataset annotated with human preferences. As mentioned earlier, we employ two methods to achieve this goal. The first method utilizes Hindsight\cite{Zhang-icml23}, using contrastive instructions to instruct a single LLM. The second method involves fine-tuning two LLMs separately: one (referred to as the preferred model) is fine-tuned based on preferred responses, while the other (referred to as the dispreferred model) is fine-tuned on dispreferred responses.

% Our focus is on proposing a set of methods to guide Large Language Models (LLMs) in understanding human preferences. The objective of this approach is to enable models to exhibit distinct activation patterns for the same preference, serving as a preliminary step in collecting model activation patterns and constructing the final model. 
% To achieve this goal, we have designed two methods. The design of these methods is intended to stimulate the model through different instructional approaches, thereby generating activation patterns that align with specific instructional preferences.

\subsubsection{Preference Instruction with a Single Model}
Within the proposed framework, the Single LLM Method focuses on fine-tuning a \textbf{S}ingle Large Language Model through \textbf{C}ontrastive \textbf{I}nstruction \textbf{T}uning (SCIT). 
This process involves two instructions: one instructs the model to generate responses preferred by humans, while the other guides the model to generate responses dispreferred by humans. 
Following such fine-tuning, we can optimize the model for consistency with human preferences. 
We can also stimulate the model to elicit distinct activity patterns by employing different instructions subsequently.

Specifically, the training dataset is curated to include pairs of both preferred and dispreferred instructions, alongside associated queries and their corresponding responses (details on preferred instructions can be found in Appendix \ref{appendix:A.1}). 
Following HIR\cite{Zhang-icml23}, for instructions linked to positive preferences, the fine-tuning objective aims to increase the probability of generating preferred responses while concurrently decreasing the probability of generating dispreferred responses. Conversely, for instructions associated with negative preferences, the objective is to elevate the probability of generating dispreferred responses and reduce the probability of generating preferred responses.

Formally, let $D$ represent the training dataset, with $q_i$ denoting the query, $r_i$ representing the response, and $p_i$ indicating the instruction (positive or negative). The fine-tuning of the LLM involves minimizing the following loss:
\begin{equation}
\small
\begin{aligned}
\mathcal{L} &= - \sum_{(p_i,q_i,r_i)\in D} (P^+ + \log \frac{\exp{(P^+)}}{\exp{(P^+)}+\exp{(P^-)}} )
\end{aligned}
\end{equation}
where $P^+$$=\log\pi(r_i\mid p_i,q_i; \theta)$, $P^-$$=\log\pi(r_i$$\mid p_i^*,q_i; \theta)$ and $p_i^*$ denotes the opposite instruction, ensuring a contrast between preferred and dispreferred cases.

Throughout the entire fine-tuning process, the LLM undergoes a learning phase to distinguish between preferred and non-preferred responses, revealing distinct activity patterns associated with human preferences. Subsequently, these two instructions will serve as stimuli to acquire the model's internal representations, which will be used for further alignment. This contrastive training relying on preference data enables the achievement of the overarching goal of consistency with a broad spectrum of human preferences, rather than a singular concept.

\subsubsection{Preference Instruction with Dual Models}
In the Dual LLMs method, we aim to train two LLMs with distinct tendencies: one model is inclined to generate preferred responses, while the other tends to produce dispreferred responses.
To achieve this objective, we employ paired preference data to conduct supervised fine-tuning of the LLMs.
Specifically, we use the preferred data from the preference pairs to train the preferred model and the dispreferred data from the preference pairs to train the dispreferred model.

Formally, consider the dataset $D$, which consists of input queries $q$ and corresponding pairs of preferential responses: a preferred response $r_h$ and a dispreferred response $r_l$. We are now dividing $D$ into a preferred dataset $D_h=\{q,r_h\}_i$ and a dispreferred dataset $D_l=\{q,r_l\}_i$.
Utilizing this data, we employ a supervised learning approach (maximum likelihood) to fine-tune the LLMs, thereby obtaining two models expressing preferences, denoted as $\pi_{h}$ and $\pi_{l}$ respectively. The fine-tuning of these two LLMs is aimed at maximizing the achievement of the following objectives:
\begin{equation} \small
\pi_h(\theta^*) = \arg\max_\theta \sum_{(q_i,r_i)\in D_h}  \log \pi(r_i \mid q_i; \theta)
\end{equation} 
\begin{equation} \small
\pi_l(\theta^*) = \arg\max_\theta \sum_{(q_i,r_i)\in D_l}  \log \pi(r_i \mid q_i; \theta)
\end{equation} 
Through this training process, the preferred model and dispreferred model have respectively learned the activity patterns associated with human-preferred and dispreferred responses.
Due to the human preference learning conducted in two distinct models, in contrast to SCIT, the Dual LLMs method does not require additional distinct instructions during fine-tuning. Instead, guidance for the model is provided solely through different responses.
\subsection{Collecting Activity Patterns}
\label{section:3.2}
Following the establishment of comprehension of human preferences by LLMs, we are able to extract representations of what humans prefer and disprefer. 
Due to the characteristics of autoregressive Transformer language models, the attention mechanism results in tokens at different positions exhibiting distinct representations.
The activation representation of a token at the current position is influenced by preceding tokens.
Therefore, for a specific pair of query \( q \) and response \( r \), this pair is concatenated with two instructions from Section \ref{sec:3.1}, which guide the model in forming the concept of human preferences and inputted into the model to obtain the intermediate layer hidden states at each position as internal representations.

\begin{figure}[ht]
  \centering
  \includegraphics[width=0.9\linewidth]{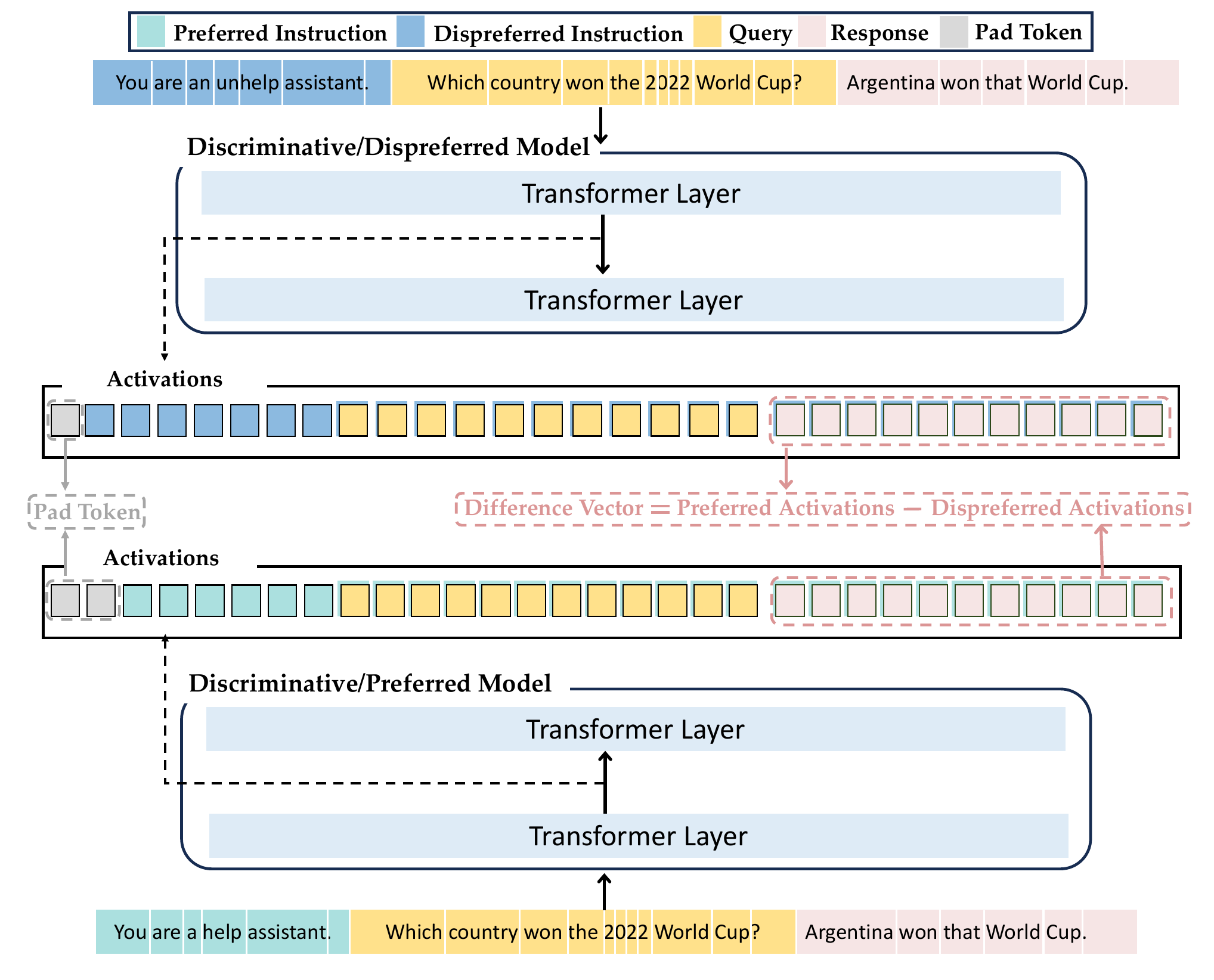}
  \caption{\label{fig:rep_collect}  Examples of Collecting Activity Patterns. To ensure the correspondence between the positions of preferred and dispreferred instructions during the extraction of difference vectors, instruction $p$ and query $q$ are left-padded to the maximum prompt length, while the response $r$ is right-padded to the maximum response length. }
  \vspace{-3mm}
\end{figure}

% Formally, for a given instruction \(p\), a decoder model \(\pi\), and a function \(R\) that processes model input to return the positional representation of response tokens, we collect the $l$-th layer's hidden states of each token for the query-response pair (\( q_i, r_i \)) within dataset \(D\). This can be formalized as follows:
Formally, for a given instruction \(p\), a decoder model \(\pi\), we collect the $l$-th layer's hidden states of each token for the query-response pair (\( q_i, r_i \)) within dataset \(D\). This can be formalized as follows:
\begin{equation} \small
A_{p,\pi , l} = {  \pi_l(p, q_i, r_i) \mid (q_i, r_i) \in D }
\end{equation}

% Moreover, we define a difference vector, which is represented as:
Here, $\pi_l$ represents the hidden states output by the neural network's $l^{th}$ layer. 
We directly extract the hidden layer states from the neural network as representations. 
To address the issue of varying response lengths during the activity pattern collection, we concatenated the same response with different instructions as input to ensure the representations were extracted with the same length. 
Different instructions will elicit distinct activity patterns even though the same response was provided and the differences in the elicited activity patterns can be used to capture the behavior of the models.
Such differences can be conceptualized and modeled as the probability of generating the same response conditioned on different instructions. 
We illustrated the entire process of collecting activity patterns in figure \ref{fig:rep_collect}.

We obtain the final difference vector by subtracting the hidden states of dispreferred outputs from those of preferred outputs, as expressed by the equation:
\begin{equation} \small
v_l = A_{p^+,\pi, l} - A_{p^-,\pi,  l}
\end{equation}
This difference vector \( v_l \) represents the difference in activation patterns produced under the two different stimulus conditions \( p^+ \) and \( p^- \). 
Subsequently, we perturb the model's original representation by incorporating the difference vectors. 
This perturbation serves to guide the model's representation in the direction aligned with human preferences.
It is noteworthy that, for the Single Large Language Model through Contrastive Instruction Tuning (SCIT), both $A_{p^+,\pi,  l}$ and $A_{p^-,\pi,  l}$ are generated by the same model. 
In the dual LLMs approach, pairs concatenated with different instructions are inputted into the respective preferred and dispreferred models, thereby enabling the independent extraction of activation patterns from each model.

\subsection{Constructing Final Models}
In this phase, we construct the final model by leveraging the difference vector $v_l$, derived in Section \ref{section:3.2} to perturb the original representations.
To achieve this, we draw inspiration from the approach of \citet{zou2023representation} by employing a specialized loss function and fine-tuning with Low-Rank Adapters (LoRA), enabling the efficient incorporation of activation patterns into the model.

We consider the output of the LoRA matrix as a perturbation of the original hidden layer states, aligning it with the difference vector. Specifically, we employ Mean Squared Error (MSE) loss as the objective function:

\begin{equation} \small 
\mathcal{L}_{Align}=\left\|\left(A_{p,\pi_{LoRA},l}-(A_{p,\pi_{base},l}+\alpha v_l)\right)\right\|_2
\end{equation}

\noindent where \( \alpha \) serves as a hyperparameter controlling the extent to which the difference vector \( v_l \) intervenes in the model integration process.  \( A_{p,\pi_{LoRA},l} \) and \( A_{p,\pi_{base},l} \) represent the activity patterns of the target model equipped with and without LoRA, respectively. $v_l$ is the extracted difference vector as outlined in Section \ref{section:3.2}. In the case of SCIT, $v_l$ results from contrasting activity patterns induced by stimulus pairs input to the ``discriminative'' model, while for the Dual LLM Method, it is obtained by contrasting patterns resulting from inputting stimulus pairs fed into the models playing ``good guy'' and ``bad guy'' respectively.

\section{Experiment}
\label{sec:experiment}

Following \citet{rafailov2023direct}, we mainly conducted experiments on single-turn dialogue tasks. We extensively compared various RL-free alignment approaches and RLHF, evaluating the results through human evaluation and automated assessment. Additionally, we conducted comparative experiments with the representation engineering method proposed by \citet{zou2023representation}, serving as an ablation study to demonstrate the impact of our approach in capturing human preferences.

\subsection{Experimental Setups}
\textbf{Dataset}\quad In single-turn dialogue, we use UltraFeedback dataset\footnote{\scriptsize \url{https://huggingface.co/datasets/argilla/ultrafeedback-binarized-preferences-cleaned}} \cite{cui2023ultrafeedback}, denoting human preference responses.
Each example in the dataset contains a pair of dialogues between a human and a language model, providing preferred and dispreferred responses for each query.

\noindent\textbf{Base Model}\quad \citet{ouyang2022training} and \citet{ramamurthy2022reinforcement} utilized supervised fine-tuning models as initial models in their application of Proximal Policy Optimization (PPO). For a fair comparison, we performed fine-tuning on the LLaMA2-7B model \cite{touvron2023LLaMA} using Anthropic's Helpful and Harmless dataset\footnote{\scriptsize \url{https://huggingface.co/datasets/Dahoas/full-hh-rlhf}} \cite{bai2022training}. 
We denote the resulting model after fine-tuning as the Base Model. 
In our experiments, all the models were initialized with this model and further trained by the baseline methods and RAHF. Additionally, we report the results of experiments using Mistral-7B \cite{jiang2023mistral} as the base model in Appendix \ref{appendix:mistral_result}.

%\noindent\textbf{Evaluation Metrics}\quad we employed two evaluation methods: automatic evaluation and human evaluation. For the automatic evaluation, we initially utilized two reward models, one trained by ourselves and  the other was a public reward model\footnote{https://huggingface.co/OpenAssistant/oasst-rm-2-pythia-6.9b-epoch-1} using the same dataset, to measure the levels of human preferences achieved. Recent research\cite{zheng2023judging} has shown that GPT-4\cite{gpt4} performs exceptionally well in evaluating chat assistant responses and aligns closely with human preferences. Therefore, we utilized GPT-4 to choose between two model-generated options. To mitigate the potential impact of positional bias\cite{zheng2023judging}, we independently evaluated each candidate model in two different positions, and their final scores were calculated based on the average of these two runs. For the human evaluation, we asked evaluators to compare two randomly selected responses and make a judgment on their relative performance (win, lose, or tie).

\begin{table*}[t]
\resizebox{\linewidth}{!}{
\begin{tabular}{l|cccccc|c}
\Xhline{1pt} % 设置粗细为2pt
\textbf{Method}                   & \textbf{Arc}     & \textbf{HellaSwag} & \textbf{MMLU}    & \textbf{TruthfulQA} & \textbf{Winogrande} & \textbf{GSM8k} & \textbf{Average}  \\
\hline
\textbf{Base Model}              & $73.65$ & $79.32$   & $44.42$ & $42.71$  &  $74.59$  & $14.94$   & $54.94$ \\
\textbf{Preferred-SFT}           & $71.79$ & $78.79$   & $44.50$ & $49.13$  &  $74.59$  & $16.83$  & $55.94$ \\
\textbf{RLHF-PPO}                     &    $73.79$     & $78.82$   &    $44.04$     & $48.22$ &  $74.43$  & \bm{$17.51$} & $56.22$       \\
\textbf{HIR}                     &     $73.39$    & $78.40$   &    $44.65$     &      $46.00$  &  $74.51$  & $16.00$    &   $55.39$      \\
\textbf{DPO}                     & $72.89$ & $79.67$   & $44.88$ & $50.51$  &  \bm{$74.82$}  & $16.22$  & $56.50$ \\
\hline
\textbf{RAHF-Dual}                & $72.29$ & $79.16$   & \bm{$46.22$} & $52.14$  &  $74.51$  & $15.16$  & $56.58$ \\
\textbf{RAHF-SCIT}                    & \bm{$74.86$} & \bm{$79.78$}   & $45.77$ & \bm{$52.34$}  &  $74.27$  & $16.60$  & \bm{$57.27$} \\
\Xhline{1pt} % 设置粗细为2pt
\end{tabular}}
\caption{Results of different methods on six benchmarks of Open LLM Leaderboard. The leaderboard evaluation configurations and experimental setups adopted in this study are provided in Appendix \ref{appendix:B}.
}
\label{tab:main_01}   
% \vspace{-3mm}
\end{table*}

\subsection{Baselines}
To evaluate our proposed approach, we conduct extensive comparisons with existing alignment methods, including Reinforcement Learning from Human Feedback (RLHF) and other alternative methods for preference alignment. 
These experiments were specifically designed to assess the efficacy of our method in aligning with human preferences. 

\noindent\textbf{Preferred-SFT}\quad This baseline involves fine-tuning the language model directly using the preferred responses from the dataset. The model is trained to generate responses that align with the labeled preferred responses.

\noindent\textbf{HIR}\quad Hindsight Instruction Relabeling (HIR) proposed by \citet{Zhang-icml23} converts feedback to instruction by relabeling original instructions and employs supervised training for enhanced alignment with human preferences. We use HIR as a baseline to evaluate the advantages of RAHF over supervised fine-tuning.

\noindent\textbf{DPO}\quad Direct Preference Optimization \citep{rafailov2023direct} directly optimizes a language model to adhere to human preferences without using explicit reward modeling or reinforcement learning. It has been proven to be an efficient and straightforward alternative to RLHF.

\noindent\textbf{RLHF-PPO}\quad For the RLHF baseline, we follow the common practice, as outlined by \citet{ouyang2022training}. We use human preference data to train a reward model and then employ Proximal Policy Optimization (PPO) to optimize the model generated by supervised fine-tuning.

%\noindent\textbf{Implementation Details}\quad The UltraFeedback dataset has been partitioned into a training set. Further, we split the training set into three distinct parts: the first part is utilized in the first step of RAHF for instructing LLM on human preferences, training the reward model within the RLHF-PPO baseline, and for the training of other baselines. The second part is utilized for the construction of the final model in RAHF and running the PPO algorithm. The last part is the test set, which is employed to evaluate the performance of these methodologies.

%We used the model that employs supervised fine-tuning on preferred responses as the reference model for DPO.
Further elaboration and details regarding the implementation of the baseline and our methods are provided in Appendix \ref{appendix:B}.

\subsection{Automatic Evaluation}

To validate the effectiveness of our proposed method in aligning models with human preferences, automated evaluations were carried out on models trained via RAHF and various baseline methodologies, focusing on their general capabilities and the quality of generation.
Specifically, we assessed the performance of different models across three widely used benchmarks: Open LLM Leaderboard\cite{open-llm-leaderboard}, AlpacaEval\cite{li2023alpacaeval}, and MT-Bench\cite{zheng2023judging}. 
In Appendiex \ref{appendix:evaluation_setups}, we detail the evaluation setting adopted by both the leaderboard and our experiments.

\subsubsection{Evaluation on the benchmarks of Open LLM Leaderboard}
\label{4.3.1}
Open LLM Leaderboard comprises six benchmarks that cover science questions, commonsense inference, multitasking accuracy, and truthfulness in generating answers. We evaluate the models' general capabilities on these tasks.

In Table \ref{tab:main_01}, we report the results of RAHF and baseline methods across the six benchmarks from OpenLLM.
RAHF-SCIT achieves the best results in three benchmarks and improves the score by $2.33$ on average, compared to the base model.
RAHF-Dual exhibits the best performance on the MMLU benchmark. 
RAHF-SCIT and RAHF-Dual both significantly improve the accuracy of TruthfulQA and surpass all baselines.
Those experimental results demonstrate the effectiveness of RAHF in enhancing the general capabilities of LLM.

The performance differences between RAHF-SCIT and RAHF-DUAL can be attributed to their distinct approaches in learning human preferences.
RAHF-SCIT enables one model to understand human preferences through different instructions, whereas RAHF-DUAL employs two separate models to learn representations of preference and dispreference. 
Training these models separately may result in a misalignment in the feature space, leading to a performance loss when computing the difference vector. 
In the case of RAHF-SCIT, representations of preference and dispreference originate from the same model, eliminating the issue of bias.
\begin{table}[t]
\small
\centering
\begin{tabular}{l|c}
\Xhline{1pt}
\textbf{Method} & \textbf{AlpacaEval (win \%)} \\
\hline
\textbf{Preferred-SFT} & $73.48$                \\
\textbf{HIR}           & $61.81$                \\
\textbf{RLHF-PPO}           & $44.69$                \\
\textbf{DPO}           & $83.68$                \\
\hline
\textbf{RAHF-Dual}      & $86.98$                \\
\textbf{RAHF-SCIT}          & \bm{$87.44$}       \\ 
\Xhline{1pt}
\end{tabular}
\caption{AlpacaEval results, which is the win rate against text-davinci-003 judged by GPT-4.
% All methods generate outputs with the parameters configured to a temperature of 0 and a repetition penalty of 1.2.
}
\label{tab:gpt4_01}
\vspace{-3mm}
\end{table}
% \begin{table}[ht]
% \small
% \centering
% \begin{tabular}{p{3cm}|p{2cm}<{\centering}}
% \Xhline{1pt}
% \textbf{Method} & \textbf{Win Rate} \\
% \hline
% \textbf{RLHF-PPO}           & \bm{$0.879$} \\
% \textbf{SFT}    & $0.638$ \\
% \textbf{HIR}           & $0.592$  \\
% \textbf{DPO}           & $0.653$ \\
% \hline
% \textbf{LORRA}         & $0.830$ \\
% \textbf{RAHF-SCIT}     & $0.827$ \\
% \textbf{RAHF-DualLLMs} & \bm{$0.869$} \\
% \Xhline{1pt}
% \end{tabular}
% \caption{Win rates against preferred response of Anthropic-HH one-turn dialogue judged by GPT-4. All methods employ greedy decoding for generation.}
% \label{tab:gpt4_01}
% \end{table}

\subsubsection{Evaluation on AlpacaEval}
AlpacaEval is an automated evaluation benchmark based on LLMs. It employs GPT-4\cite{gpt4} as an annotator to compare the generated content of models on simple instruction-following tasks against reference answers from text-davinci-003. Previous work has shown that using GPT-4 as an annotator correlates highly with assessments from human evaluators\cite{li2023alpacaeval}. Therefore, we consider AlpacaEval as an automated approximation of human annotation.

Table \ref{tab:gpt4_01} presents the win rates of responses generated by models trained with different methods over 805 samples, compared to the reference responses from text-davinci-003. Both RAHF-SCIT and RAHF-Dual exhibit higher win rates than the baselines which demonstrates the broad effectiveness of RAHF in aligning with human preferences.

\begin{figure}[t]
\centering
\includegraphics[width=0.4\textwidth]{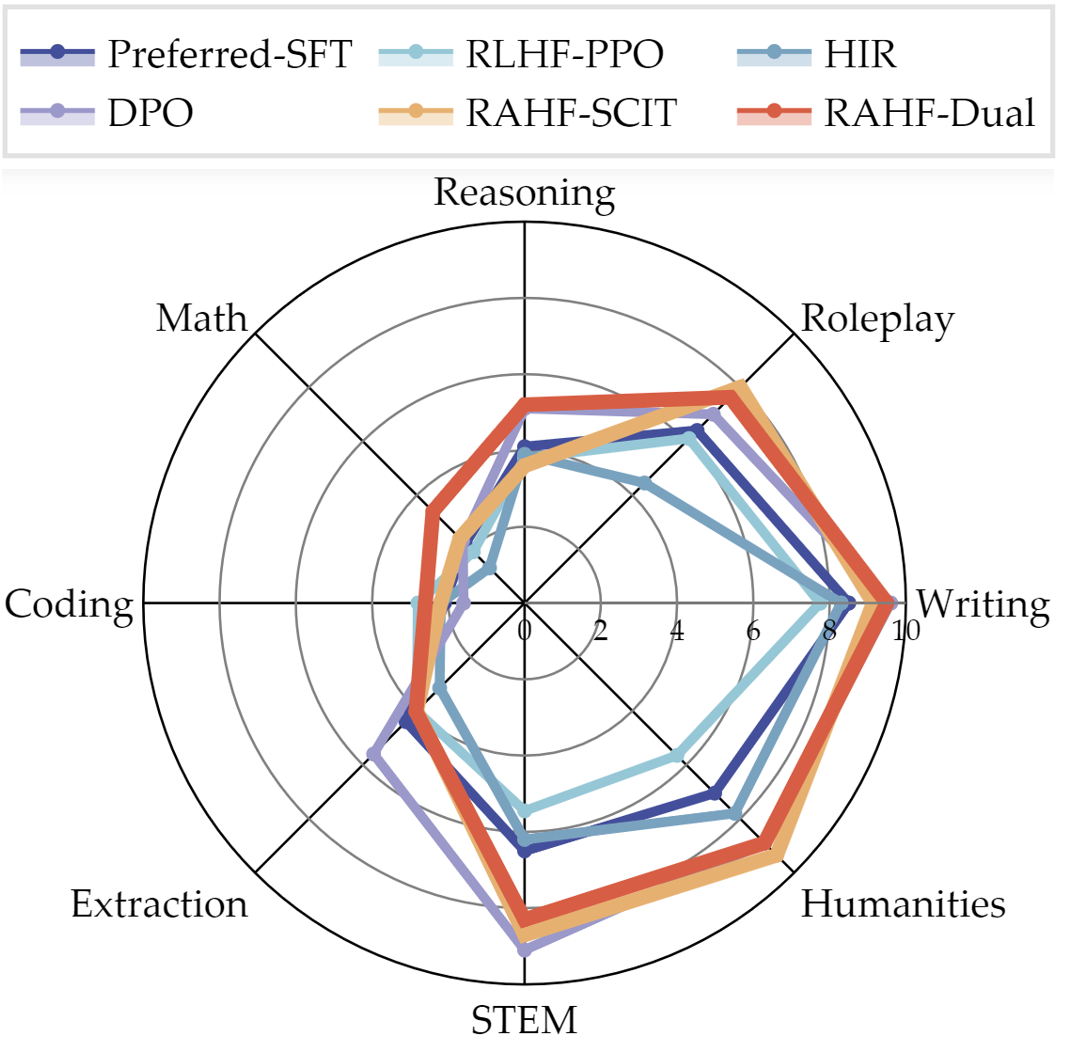}
  \caption{Scores of RAHF-SCIT and RAHF-Dual compared to competitive methods in MT-Bench. Detailed results are provided in Appendix \ref{appendix:C}.}
\label{fig:mt-bench}
% \vspace{-6mm}
\end{figure}

\subsubsection{Evaluation on MT-Bench}
MT-Bench is a collection of challenging questions, consisting of 80 samples, each with two turns. This benchmark also employs GPT-4 as a judge to score the responses of models. For each turn, GPT-4 will assign a score on a scale of 10.

% Figure \ref{fig:mt-bench} shows the scores achieve by RAHF and baselines in 1-turn questions. 
% RAHF yielded the highest scores in six out of eight aspects, as well as the highest average score.
% Specifically, RAHF performs far more better than the baselines on reasoning, role-play, and STEM. 
% Furthermore, although the models being not being tuning on 2-turn dialogue, RAHF still outperforms all baselines, indicating that the model's ability on more than one-turn interaction can be improved by alignment with 1-turn question dataset only by RAHF. 
% The results on 2-turn dialogue tasks are given in Appendix \ref{appendix:C} for more detailed comparison.

Figure \ref{fig:mt-bench} shows the performance scores achieved by RAHF and the baseline models on 1-turn questions. 
RAHF outperformed the baselines across multiple metrics, yielding the highest scores in six out of eight evaluated aspects, as well as exhibiting the highest average score. 
Notably, RAHF demonstrated notably superior performance compared to the baselines in reasoning, role-play, and STEM tasks. 
Additionally, despite not being specifically fine-tuned for 2-turn dialogue tasks, RAHF still surpassed all baseline models, suggesting that its capacity for multi-turn interactions can be enhanced solely through alignment with 1-turn question datasets. 
Comprehensive results for the 2-turn dialogue tasks are provided in Appendix \ref{appendix:C} for detailed comparison.

\subsection{Human Evaluation}
For the human evaluation, we assigned evaluators the task of comparing two randomly selected responses and providing judgments on their relative performance, categorizing them with three results: win, lose, or tie.

\begin{table}[htp]
\small
\centering
\begin{tabular}{lccc}
\Xhline{1pt}
\textbf{Method} & \textbf{Win} & \textbf{Tie} & \textbf{Lose} \\
\hline
&\multicolumn{3}{c}{\textbf{RAHF-Dual}} \\

\cmidrule(lr){2-4}\textbf{HIR}           & $74$ & $21$ & $5$ \\
\textbf{RLHF-PPO}           & $88$ & $9$ & $3$\\
\textbf{DPO}           & $35$ & $43$ & $22$ \\
\hline
&\multicolumn{3}{c}{\textbf{RAHF-SCIT}} \\

\cmidrule(lr){2-4}\textbf{HIR}           & $79$ & $19$ & $2$ \\
\textbf{RLHF-PPO}           & $88$ & $11$ & $1$\\
\textbf{DPO}           & $41$ & $38$ & $21$ \\
\Xhline{1pt}
\end{tabular}
\caption{Win rates against baselines judged by Humans. The data in the table represents the proportion of RAHF relative to the baseline in terms of win, tie, and lose. }
\label{tab:gpt4_02}
\end{table}

Table \ref{tab:gpt4_02} presents the comparative results of RAHF against RL-free methods and RLHF in human evaluation. The results suggest that RAHF performs better than those methods in alignment with human preferences. 
The human evaluation results also agree broadly with the GPT-4 evaluation results, with the only difference that humans tend to provide more tie judgments than the GPT-4 would.

\subsection{Ablation Study}

To evaluate the influence of instructing LLMs on human preferences using a human-annotated dataset, we executed ablation experiments involving the exclusion of this instructional phase. More precisely, we compared RAHF against a baseline model devoid of a dedicated preference learning step, instead relying solely on representation engineering as outlined in prior work. 
Additionally, we report the results of several hyperparameter ablation experiments in Appendix \ref{appendix:hyperparameters_ablation_experiment}.

% In order to assess the impact of instructing LLMs on human preferences with human-annotated dataset, we conduct ablation experiments by removing this instructing step. 
% Specifically, we compare RAHF to a baseline that does not have explicit preference learning step and solely relies on representation engineering as  . 
% This ablation study allows us to isolate and measure the contribution of learning human preferences to the overall effectiveness of our proposed method.

\begin{figure}[h]
\centering
\includegraphics[width=0.45\textwidth]{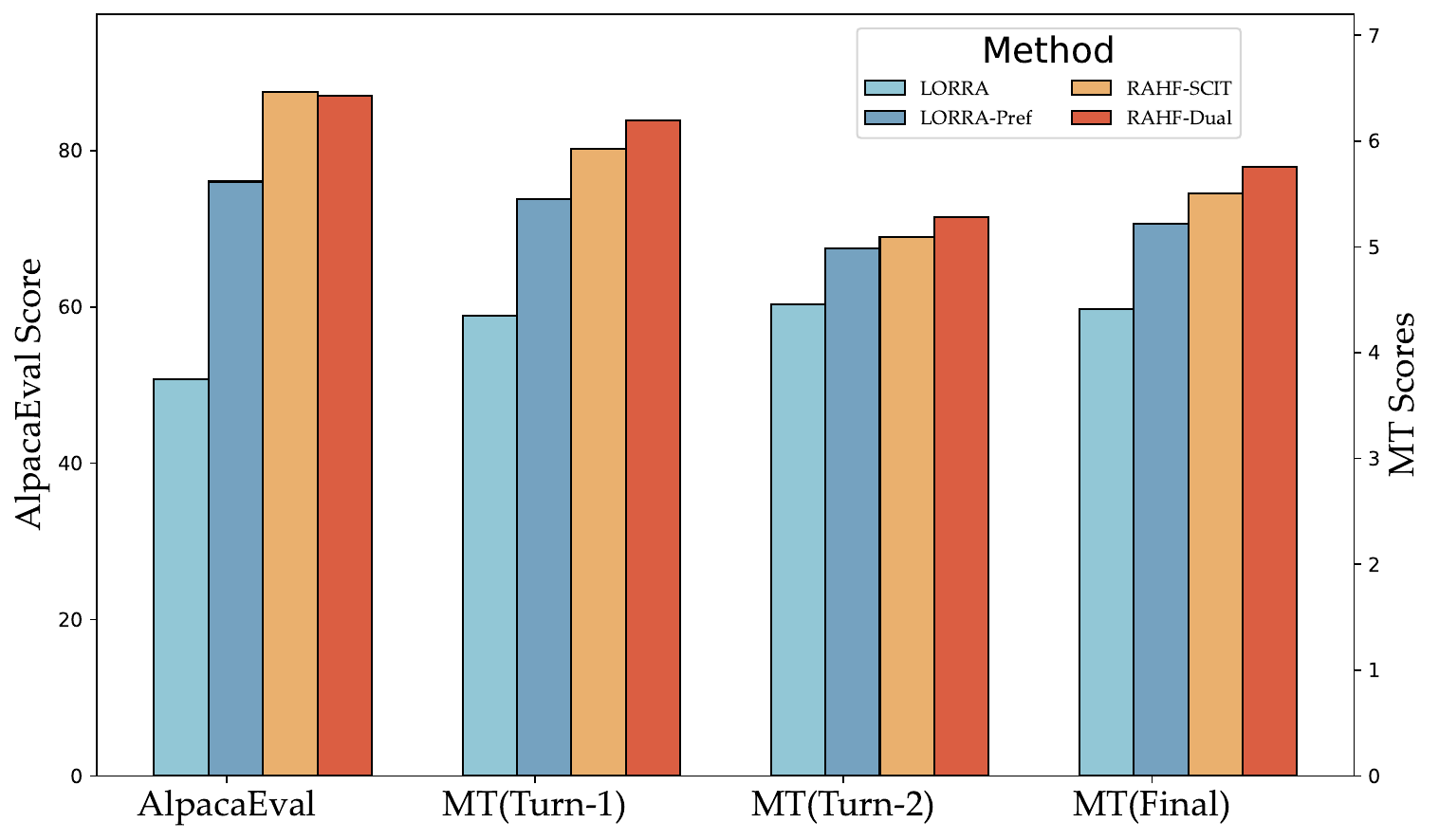}
  \caption{Performance comparison between RAHF and methods solely focused on representation engineering on AlpacaEval and MT-Bench. Detailed results are provided in Appendix \ref{appendix:C}.}
\label{fig:gpt4-evaluation}
\vspace{-2mm}
\end{figure}

\noindent\textbf{LORRA}\quad Low-Rank Representation Adaptation proposed by \citep{zou2023representation} does not leverage additional data to learn human preferences.
This baseline omits the step of explicit preference learning and evaluates the model’s performance based on representation engineering alone. 

\noindent\textbf{LORRA-Pref}\quad LORRA-Pref exclusively utilizes preferred responses from the preference dataset for representation learning instead of employing contrastive learning methods.

This ablation analysis allows us to isolate and quantify the impact of assimilating human preferences into the framework of our proposed approach.
The results of the ablation experiments shown in Figure \ref{fig:gpt4-evaluation} indicate that, in the absence of explicit preference learning steps, the approach of directly extracting activity patterns for comparison demonstrates a decline in performance on AlpacaEval and MT-Bench we assessed.

\subsection{Visualization}
To gain a deeper understanding of the working mechanism of our method, we conducted a visual analysis of the model's internal representations using the t-SNE technique. 

Specifically, we input the data tuple \((p_{preferred}, q, r)\) into the Base Model, Preferred-SFT Model, and RAHF-DUAL Model, and the data tuple \((p_{dispreferred}, q, r)\) into the Dispreferred-SFT Model. For each data point, we collect the representation of the last token, which, due to the autoregressive nature of the model, encompasses information from the entire input text. Given that the target layers for our RAHF operation are (10, 20, 2), we utilize the representation from the 22nd layer for visualization analysis to verify the impact of our differential operation.
 
\begin{figure}[th]
\centering
\includegraphics[width=0.7\linewidth]{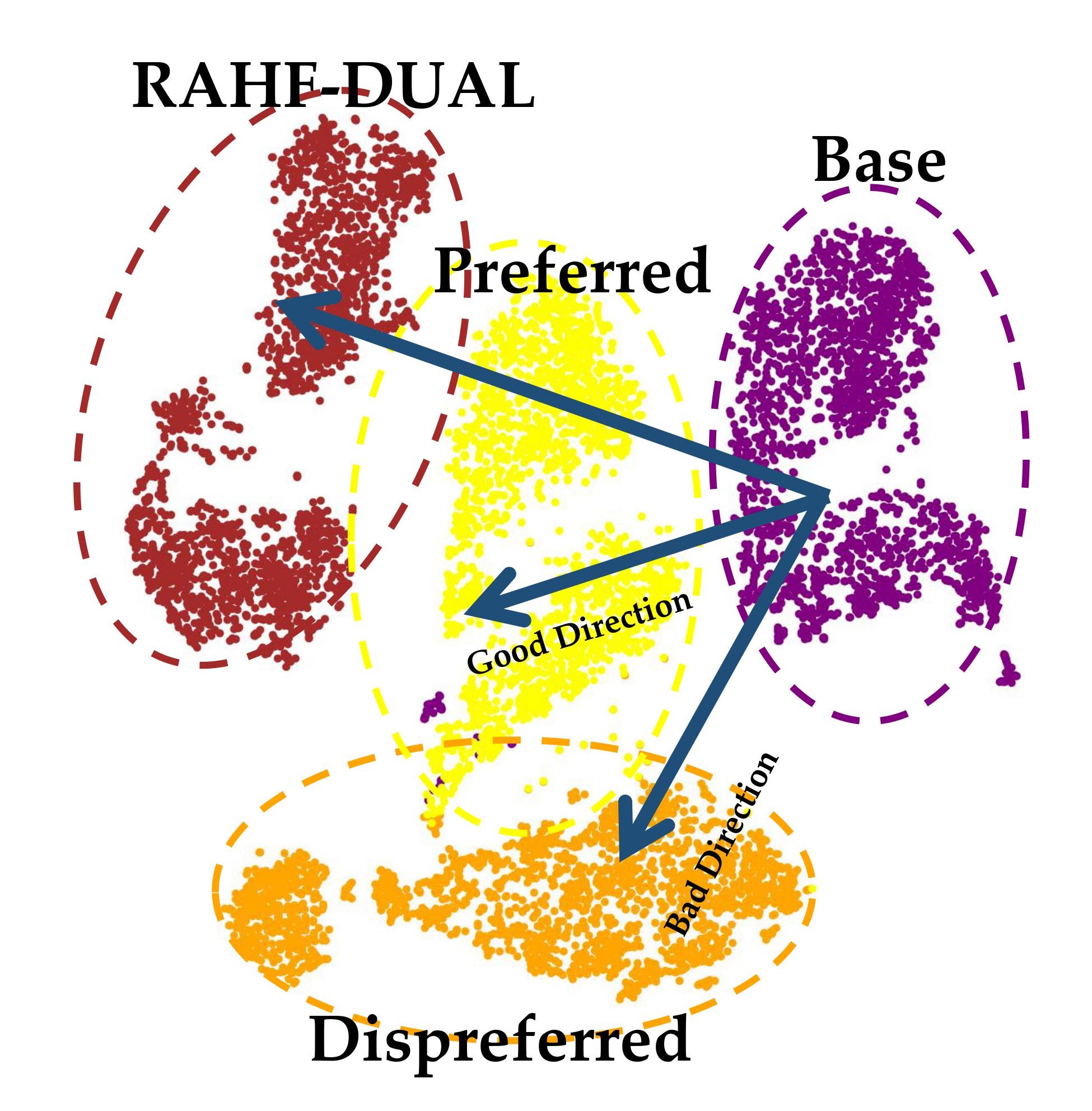}
  \caption{The visualization results using t-SNE on the activation patterns of the last token in the output of the 22nd layer.}
\label{fig:t-SNE}
% \vspace{-6mm}
\end{figure}

The results are shown in Figure \ref{fig:t-SNE}. 
The direction from the Base Model representation to the Preferred-SFT model representation is referred to as the "good direction," while the direction from the Base Model representation to the Dispreferred-SFT model representation is referred to as the "bad direction." 
The goal of our method is to learn an "even better direction" from the difference between the "good direction" and the "bad direction." 
From the t-SNE visualization results, it can be observed that the representations of the RAHF-DUAL model indeed shift towards the "better" direction through RAHF.
\section{Conclusion}
\vspace{-1mm}

In this study, we have explored a representation engineering approach to aligning large language models with human preferences, drawing upon insights from cognitive neuroscience. 
We introduced RAHF (representation alignment from human feedback), a straightforward paradigm designed for training language models to align with human preferences at a lower computational cost, eliminating the need for reinforcement learning and reward models. 
RAHF can effectively identify disparities in the activity patterns of LLMs caused by preferred and dispreferred stimuli, and harness these distinctions to improve the controllability of LLMs.
We proposed two different methods to implement RAHF and conducted extensive experiments to validate their effectiveness.
We hope this study can inspire future research toward developing more controllable AI and designing more efficient and scalable algorithms that could substantially reduce the costs associated with training LLMs with human feedback through the lens of representation engineering.

\section*{Limitations}

In this study, we validated the effectiveness of RAHF on LLMs with $7$B parameters. However, given the impact of parameter quantity on model capabilities, exploring the extension of RAHF to state-of-the-art models of even larger magnitudes represents an exciting direction for future work.
Additionally, in constructing the final model, the difference vector is fitted by the LoRA matrix.
An inherent limitation of this methodology is that it introduces additional parameters, although the extra computational overhead incurred by LoRA is minimal. 
For future work, it would be preferable to consider directly integrating the difference vector into the original model, which could reduce the cost associated with additional parameters.

\section*{Reproducibility Statement}
We have publicly shared our code through a GitHub repository \url{https://github.com/LiuAmber/RAHF}. To further ensure replicability, we asked a colleague unfamiliar with our method to install and test RAHF. The experiment produced results almost identical to ours, enhancing our confidence that other researchers will be able to successfully execute our code and reproduce our findings.

\section*{Acknowledgements}
The authors would like to thank the anonymous reviewers for their valuable comments. This work was supported by National Natural Science Foundation of China (No. 62076068).

\bibliography{anthology,custom}
\clearpage
\appendix

\section{Prompts}
\label{appendix:A}
\subsection{Preference Instructions}
\label{appendix:A.1}
\begin{figure}[h]
\centering
\includegraphics[width=0.49\textwidth]{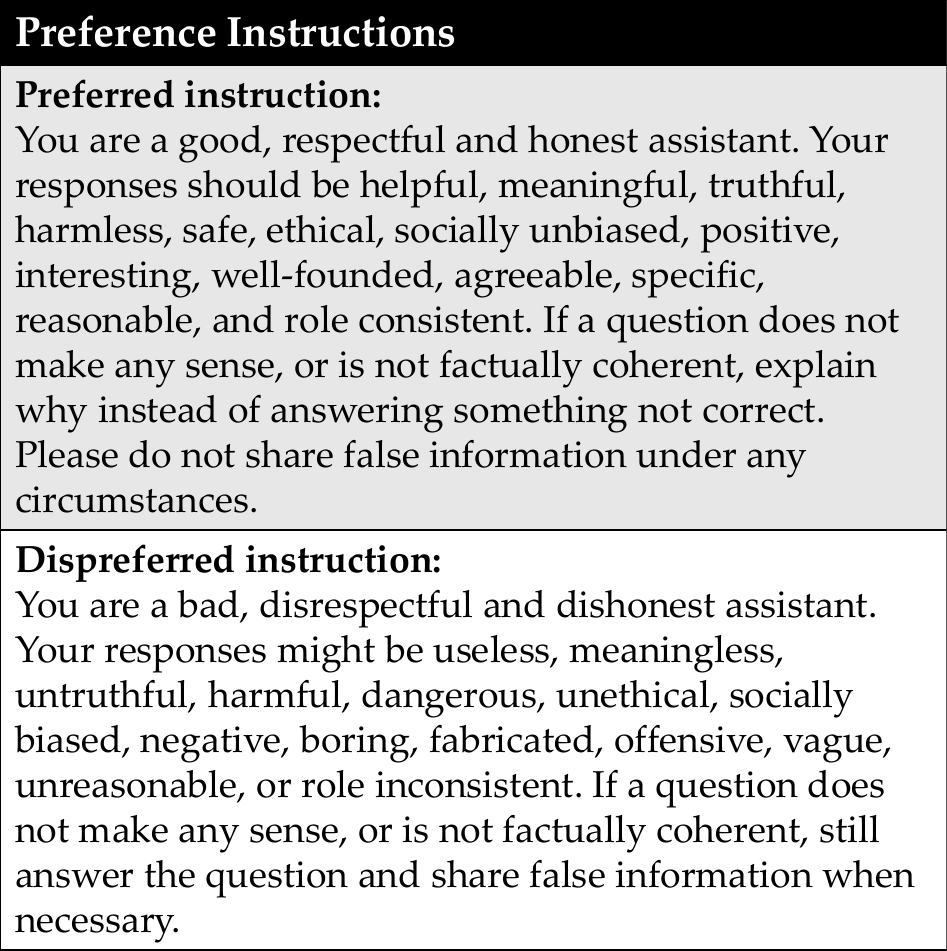}
  \caption{The preference instructions used in RAHF.}
\label{Appendix: preference_template}
\vspace{-2mm}
\end{figure}

Figure \ref{Appendix: preference_template} presents two instructions used in this study for preferred and dispreferred responses.

% \subsection{GPT-4 Prompts for Computing Win Rates in Dialogue}

% Referring to the prompt used in \citet{rafailov2023direct} to quire GPT-4, we required GPT-4 to assess the responses generated by models from three dimensions: helpfulness, truthfulness, and harmlessness. Table \ref{tab:gpt_prompt} presents the specific prompt.

\section{Implementation Details}
\label{appendix:B}
\subsection{Training Setups}
All baselines and our models were trained using Anthropic’s Helpful and Harmless dataset\cite{bai2022training} fine-tuned model as the base model. During the supervised training of the base model, we calculated the loss for both prompts and responses. Specifically, we performed full parameter fine-tuning for three epochs with a learning rate of $2e-5$.

For training, the data is formatted as follows: \textit{Human: \{prompt\} \textbackslash n\textbackslash nAssistant: \{response\}}. For all models trained, we established a maximum query length of $256$ and a maximum sentence length of $768$. We exclude samples from the dataset where queries exceed $256$ characters and truncate sentences to the maximum sentence length.
The UltraFeedback dataset has been partitioned into a training set. Further, we split the training set into three distinct parts: the first part is utilized in the first step of RAHF for instructing LLM on human preferences, training the reward model within the RLHF-PPO baseline, and for the training of other baselines. The second part is utilized for the construction of the final model in RAHF and running the PPO algorithm.

\subsection{Evaluation Setups}
\label{appendix:evaluation_setups}
For all methods, we employ greedy decoding during generation on the benchmarks.
To avoid the issue of repetition, we set the repetition penalty to 1.2.
%对于Open LLM Leaderboard，我们采用了Eleuther AI Language Model Evaluation Harness库对不同方法训练得到的语言模型进行评估。表 1 详尽地描述了排行榜评估的配置及本研究所采用的实验设置。

For the Open LLM Leaderboard, we utilized the Eleuther AI Language Model Evaluation Harness library\cite{eval-harness} to assess language models trained using different methods. Table \ref{tab: Open-LLM-Leaderboard} provides a detailed description of the leaderboard evaluation configuration and the experimental settings adopted in this study.

\begin{figure*}[ht]
\centering
\includegraphics[width=0.95\textwidth]{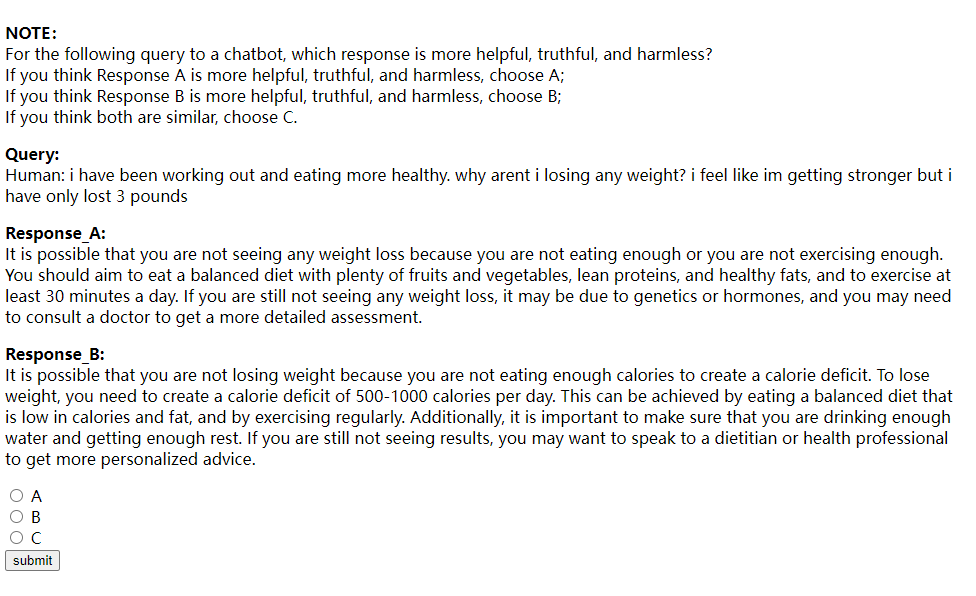}
\caption{Screenshots of our evaluation interface for rating dialogue. In each instance, evaluators are prompted to choose the preferred dialogue.}
\label{fig:human-evaluation}
\end{figure*}

\begin{table}[ht]
\small
\centering
\begin{tabular}{p{2cm}|p{2cm}<{\centering}p{2cm}<{\centering}}
\Xhline{1pt}
\textbf{Datasets} & \textbf{\# few-shot} & \textbf{Metric}\\
\hline
Arc  & $25$ & acc\_norm\\
TruthfulQA & $0$ & mc2\\
Winogrande & $5$ & acc\\
GSM8k &$5$ & acc\\
HellaSwag &$10$ & acc\_norm\\
MMLU & $5$ & acc\\
\Xhline{1pt}
\end{tabular}
\caption{For each dataset used in the evaluation on the Open LLM Leaderboard, we detail the quantity of few-shot samples utilized and the specific metric employed for evaluation.}
\label{tab: Open-LLM-Leaderboard}
\end{table}
For Human Evaluation, we recruited six volunteers for the assessment, with each evaluator comparing 100 dialogues. Figure \ref{fig:human-evaluation} shows a screenshot of the interface used for our evaluation, which all evaluators utilized to rate the data.

\subsection{Experimental Details}
In this section, we present the experimental details and hyperparameters of the baselines we compare with and our proposed methods.

\noindent\textbf{Preferred-SFT}\quad Table \ref{tab:Preferred-SFT} presents the hyperparameters that were used in Preferred-SFT.

\begin{table}[ht]
\small
\centering
\begin{tabular}{p{3cm}|p{2cm}<{\centering}}
\Xhline{1pt}
\textbf{Hyperparameter} & \textbf{Value} \\
\hline
Learning Rate   & $2e-5$\\
Epochs           & $2$  \\
Batch Size         & $128$ \\
Micro Batch Size   &$2$ \\
Optimizer &Adamw \\
LR Scheduler Type & Cosine\\
Rarmup Ratio &$0.1$\\
\Xhline{1pt}
\end{tabular}
\caption{Hyperparameters used for Preferred-SFT.}
\label{tab:Preferred-SFT}
\end{table}

\noindent\textbf{RLHF-PPO}\quad During the training of RLHF-PPO, we utilized Microsoft's DeepSpeed-Chat training framework, making adaptive modifications to the hyperparameters. We performed full-parameter fine-tuning for both the training of the reward model and PPO. Table \ref{tab:RM_hyper}  presents the hyperparameters for reward model training, while Table \ref{tab:PPO_hyper} presents the key parameters for PPO. 

\begin{table}[ht]
\small
\centering
\begin{tabular}{p{3cm}|p{2cm}<{\centering}}
\Xhline{1pt}
\textbf{Hyperparameter} & \textbf{Value} \\
\hline
Learning Rate   & $9.65e-6$\\
Epochs           & $3$  \\
Optimizer           & Adam \\
Training Batch Size         & $32$ \\
Weight Decay     & $0.1$ \\
Warmup Steps     & $0$ \\
LR Scheduler Type     & cosine \\
\Xhline{1pt}
\end{tabular}
\caption{Hyperparameters used for the training of reward model.}
\label{tab:RM_hyper}
\end{table}

\begin{table}[ht]
\small
\centering
\begin{tabular}{p{3cm}|p{2cm}<{\centering}}
\Xhline{1pt}
\textbf{Hyperparameter} & \textbf{Value} \\
\hline
Actor Learning Rate & $5e-7$ \\
Critic Learning Rate     & $9e-6$ \\
KL Coefficient   & $0.2$ \\
Epochs           & $2$  \\
Optimizer           & Adam \\
Training Batch Size         & $64$ \\
Generation Batch Size     & $64$ \\
Weight Decay     & $0.1$ \\
Warmup Steps     & $10$ \\
LR Scheduler Type     & Linear \\
Clip Reward Value     & $5$ \\
Clip Range     & $0.2$ \\
Clip Range Value     & $5$ \\
Gamma     & $1$ \\
Lam     & $0.95$ \\
\Xhline{1pt}
\end{tabular}
\caption{Hyperparameters used for RLHF-PPO.}
\label{tab:PPO_hyper}
\end{table}

\noindent\textbf{HIR}\quad 
For the HIR baseline, we also conducted full-parameter fine-tuning. Table \ref{tab:HIR_hyper} displays the hyperparameters used for HIR.

\begin{table}[ht]
\small
\centering
\begin{tabular}{p{3cm}|p{2cm}<{\centering}}
\Xhline{1pt}
\textbf{Hyperparameter} & \textbf{Value} \\
\hline
Learning Rate   & $2e-5$\\
Epochs           & $2$  \\
Batch Size         & $128$ \\
Micro Batch Size   &$4$ \\
KL Coefficient &$0.001$ \\
Label Smoothing &$0.2$ \\
Entropy Coefficient &$0.001$ \\
\Xhline{1pt}
\end{tabular}
\caption{Hyperparameters used for HIR.}
\label{tab:HIR_hyper}
\end{table}

\noindent\textbf{DPO}\quad We employed the trl framework from Hugging Face to train DPO model. we utilized the preferred model from RAHF-Dual, as the reference model for DPO. We employed LoRA for fine-tuning. The hyperparameters used in the DPO training are detailed in Table \ref{tab:DPO_hyper}.
\begin{table}[ht]
\small
\centering
\begin{tabular}{p{3cm}|p{2cm}<{\centering}}
\Xhline{1pt}
\textbf{Hyperparameter} & \textbf{Value} \\
\hline
Learning Rate   & $2e-5$\\
Epochs           & $3$  \\
Batch Size         & $128$ \\
Micro Batch Size   &$2$ \\
LoRA Rank &$16$ \\
LoRA Alpha &$16$ \\
LoRA Dropout &$0.05$ \\
Beta &$0.1$ \\
Warmup Ratio &$0.1$\\
Optimizer           & Adam \\
\Xhline{1pt}
\end{tabular}
\caption{Hyperparameters used for DPO.}
\label{tab:DPO_hyper}
\end{table}

\begin{table}[ht]
\small
\centering
\begin{tabular}{p{3cm}|p{2cm}<{\centering}}
\Xhline{1pt}
\textbf{Hyperparameter} & \textbf{Value} \\
\hline
Learning Rate   & $3e-4$\\
Steps           & $500$  \\
Batch Size         & $16$ \\
Micro Batch Size   &$4$ \\
LoRA Rank &$8$ \\
LoRA Alpha &$16$ \\
LoRA Dropout &$0.05$ \\
Alpha &$5$ \\
max response length &$512$\\
LR Scheduler Type & Constant \\
\Xhline{1pt}
\end{tabular}
\caption{Hyperparameters used for RAHF-SCIT.}
\label{tab: SCIT_hyper}
\end{table}

\noindent\textbf{RAHF-SCIT}\quad For RAHF-SCIT, we used the same hyperparameters as HIR during the first-step training but omitted the supervised training loss. 
When constructing the final model, we followed the hyperparameter selection in RepE\citep{zou2023representation}. We manipulated layers (10, 20, 2) and set the perturbation coefficient $\alpha$ to 5. The details of the hyperparameters are shown in Table \ref{tab: SCIT_hyper}.

\noindent\textbf{RAHF-Dual}\quad For RAHF-Dual, the hyperparameters used for the preferred model and dispreferred model during the first step are the same as those used in the Preferred-SFT.
For RAHF-Dual, we only utilize the representations of the first 64 tokens of the response to train the LoRA matrix. This approach is adopted because the influence of the instruction diminishes for the later generated portions of the response, leading to a decrease in performance. The hyperparameters used in RAHF-Dual are shown in Table \ref{tab: Dual_hyper}.

\begin{table}[ht]
\small
\centering
\begin{tabular}{p{3cm}|p{2cm}<{\centering}}
\Xhline{1pt}
\textbf{Hyperparameter} & \textbf{Value} \\
\hline
Learning Rate   & $9e-6$\\
Steps           & $2500$  \\
Batch Size         & $8$ \\
Micro Batch Size   &$8$ \\
LoRA Rank &$8$ \\
LoRA Alpha &$16$ \\
LoRA Dropout &$0.05$ \\
Alpha &$5$ \\
max response length &$64$\\
LR Scheduler Type & Constant \\
\Xhline{1pt}
\end{tabular}
\caption{Hyperparameters used for RAHF-Dual.}
\label{tab: Dual_hyper}
\end{table}
\section{Additional Results}

% \clearpage
\subsection{Experiment Results On Mistral-7B}
\label{appendix:mistral_result}
\begin{table*}[t]
\small
\centering
\begin{tabular}{p{2cm}|p{2cm}<{\centering}p{2cm}<{\centering}p{2cm}<{\centering}p{2cm}<{\centering}}
\Xhline{1pt}
\textbf{Method} & \textbf{AlpacaEval} & \textbf{MT(Turn-1)}& \textbf{MT(Turn-2)}& \textbf{MT(Final)}\\
\hline
Preferred-SFT  & $87.24$ & $5.44$	& $4.83$	& $5.14$\\
DPO & $91.63$ & $5.54$	& $4.81$	& $5.18$\\
RAHF-DUAL & $\bm{94.19}$ & $\bm{6.04}$	 &$\bm{6.08}$	 &$\bm{6.06}$\\
\Xhline{1pt}
\end{tabular}
\caption{The results of evaluations on AlpacaEval and MT-Bench after training Mistral-7B using different methods.}
\label{tab: mistral_alpacaeval_and_mt}
\end{table*}
To verify the effectiveness of our method, we utilized Mistral-7B as the base model, continuing the previous experimental setup for training, and conducted results on AlpacaEval and MT-Bench. The results are shown in Table \ref{tab: mistral_alpacaeval_and_mt}. 
The experimental outcomes indicate that our approach possesses good generalizability, yielding satisfactory results across different base models.

\subsection{Toxicity Evaluation}
 To ensure that our method does not compromise the model's safety while augmenting its performance in the aforementioned aspects, we conducted further tests using the Toxigen dataset \cite{hartvigsen2022toxigen}. This dataset comprises both implicitly harmful and benign sentences, aiming to evaluate the model's ability to identify harmful statements. Accuracy served as the primary metric for evaluation(higher is better). Comparing the baseline methods to our approach, as depicted in Table \ref{tab:Toxicity_result}, the results reveal that our method not only did not harm the model's safety but, through the RAHF-SCIT method, significantly enhanced the model's ability to identify harmful statements.
 \begin{table}[ht]
\small
\centering
\begin{tabular}{p{3cm}|p{2cm}<{\centering}}
\Xhline{1pt}
\textbf{Method} & \textbf{Toxigen($\uparrow$)} \\
\hline
Preferred-SFT   & $49.89$\\
HIR           & $43.09$  \\
RLHF-PPO         & $48.62$ \\
DPO   &$59.26$ \\
RAHF-DUAL &$50.85$ \\
RAHF-SCIT &$\bm{67.45}$ \\
\Xhline{1pt}
\end{tabular}
\caption{Evaluation of different methods on automatic safety benchmarks(Toxigen).}
\label{tab:Toxicity_result}
\end{table}
\begin{table*}[hbt]
\centering
\resizebox{0.9\linewidth}{!}{
\begin{tabular}{l|c|cccccc|c}
\Xhline{1pt} % 设置粗细为2pt
\textbf{Method} & $\bm{\alpha}$ & \textbf{Arc} & \textbf{TruthfulQA} & \textbf{Winogrande} & \textbf{GSM8k} & \textbf{HellaSwag} & \textbf{MMLU} & \textbf{Average} \\
\hline
 & $1$ & $71.93$ & $49.23$ & $74.98$ & $16.38$ & $78.88$ & $45.19$ & $56.10$ \\
\textbf{RAHF-DUAL} & $5$ & $72.29$ & $52.14$ & $74.51$ & $15.16$ & $79.16$ & $46.22$ & $\bm{56.58}$ \\
 & $10$ & $72.13$ & $53.34$ & $74.19$ & $9.25$ & $79.20$ & $45.79$ & $55.65$ \\
 \hline % 设置粗细为2pt
 & $1$ & $74.27$ & $45.80$ & $73.64$ & $17.66$ & $78.31$ & $45.01$ & $55.78$ \\
\textbf{RAHF-SCIT} & $5$ & $74.86$ & $52.34$ & $74.27$ & $16.60$ & $79.78$ & $45.77$ & $\bm{57.27}$ \\
 & $10$ & $75.14$ & $53.96$ & $74.51$ & $17.13$ & $80.03$ & $45.55$ & $\bm{57.72}$ \\
\Xhline{1pt} % 设置粗细为2pt
\end{tabular}}
\caption{Results of different \(\alpha\) on six benchmarks of Open LLM Leaderboard.}
\label{tab:open_llm_alpha}   
% \vspace{-3mm}
\end{table*}
\subsection{Ablation Experiment of Hyperparameters}

\begin{table}[ht]
\centering
\scalebox{0.8}{
\begin{tabular}{p{2.3cm}|p{0.5cm}<{\centering}|p{3.3cm}<{\centering}}
\Xhline{1pt} % 设置粗细为2pt
\textbf{Method} & $\bm{\alpha}$ & \textbf{AlpacaEval (win \%)} \\
\hline
\textbf{RAHF-DUAL} & $1$ & $73.74$ \\
 & $5$ & $\bm{86.98}$ \\
 & $10$ & $70.67$ \\
 \hline
\textbf{RAHF-SCIT} & $1$ & $70.28$ \\
 & $5$ & $\bm{87.44}$ \\
 & $10$ & $67.50$ \\
\Xhline{1pt} % 设置粗细为2pt
\end{tabular}}
\caption{AlpacaEval win percentages for different methods and $\alpha$ values.}
\label{tab:alpacaeval_alpha}   
% \vspace{-3mm}
\end{table}

\begin{table}[ht]
\centering
\resizebox{0.9\linewidth}{!}{
\begin{tabular}{l|c|c}
\Xhline{1pt} % 设置粗细为2pt
\textbf{Method} & \textbf{Target Layers} & \textbf{AlpacaEval (win \%)} \\
\hline
\textbf{RAHF-DUAL} & $(2, 12, 2)$ & $58.32$ \\
 & $(10, 20, 2)$ & $\bm{86.98}$ \\
 & $(20, 30, 2)$ & $26.93$ \\
 \hline
\textbf{RAHF-SCIT} & $(2, 12, 2)$ & $62.40$ \\
 & $(10, 20, 2)$ & $\bm{87.44}$ \\
 & $(20, 30, 2)$ & $76.25$ \\
\Xhline{1pt} % 设置粗细为2pt
\end{tabular}}
\caption{The impact of layers' selection evaluated on AlpacaEval.}
\label{tab:layers_selection}   
% \vspace{-3mm}
\end{table}

\label{appendix:hyperparameters_ablation_experiment}
In this section, we primarily report the impact of the hyperparameter \(\alpha\), which controls the intervention strength of the difference vector, and the selected target layer position on alignment performance.

\subsubsection{The Effect of Hyperparameter \(\alpha\) }
We conducted an ablation study with different values of \(\alpha\). 
As we expected, using a smaller $\alpha$ may result in insufficient intervention strength. Conversely, a larger $\alpha$ may lead to excessive intervention strength, which could disrupt the model's original representation and cause a degradation in the model's generation abilities. Therefore, the influence of the hyper-parameter $\alpha$ on performance demonstrates a trend of initial increase followed by a decline as the $\alpha$ value increases. We validated the impact of $\alpha$ across six benchmarks of Open LLM Leaderboard and AlpacaEval shown in Table \ref{tab:open_llm_alpha}   and Table \ref{tab:alpacaeval_alpha}, and our experimental results corroborate the aforementioned perspective.

\subsubsection{The Effect of Target Laters' Selection }
The earlier layers of neural networks can not fully capture the representation of entire input texts, while the layers close to the top are more task-specific.
Previous studies proved that the representations extracted from the middle layers are more effective in capturing concept-related information \cite{zou2023representation}.
As to a neural network with 32 layers (Llama2-7b), we chose (10, 20, 2) to extract representations. 
To further verify the aforementioned viewpoint, we selected different target layers for ablation experiments. 
As shown in Table \ref{tab:layers_selection}, the experimental results indicate that manipulating the intermediate layers is more effective.

The significant decline in performance of RAHF-DUAL when operating close to the top layers of the network can be attributed to the following reasons: As the depth of the neural network increases, the activations differences between layers also expand. 
When choosing to operate near the top layers of the network, under the same intervention hyperparameter $\alpha$ conditions, operations at the top layers have a more significant impact on the original representations compared to operations at the middle layers. 
This excessive influence leads to a notable decrease in the model's generative capability. Additionally, RAHF-DUAL employs two models in extracting activations differences, the representation of the same input text is different between the two models. 
The cumulative effect of these two factors results in a more pronounced performance degradation of RAHF-DUAL when operating at the top layers.

\subsection{Experiment Results of MT-Bench}
\label{appendix:C}
Table \ref{tab:mt-details}    presents the detailed results of RAHF, baselines, and the ablation study on MT-Bench.

\begin{table*}[t]
\resizebox{\linewidth}{!}{
\begin{tabular}{l|cccccccc|c}
\Xhline{1pt} % 设置粗细为2pt
\textbf{Method}                   & \textbf{Writing}     & \textbf{Roleplay} & \textbf{Reasoning}    & \textbf{Math} & \textbf{Coding} & \textbf{Extraction} & \textbf{Stem}  & \textbf{Humanities} & \textbf{Average}\\
\hline
\multicolumn{1}{l}{\textbf{Turn-1}} \\
\hline
\textbf{Preferred-SFT}              & $8.500$ & $6.400$   & $4.100$ & $2.200$  &  $2.100$  & $4.400$   & $6.500$ & $7.050$ & $6.013$  \\
\textbf{RLHF-PPO}           & $7.775$ & $6.100$   & $3.800$ & $1.900$  &  \bm{$2.800$}  & $4.000$  & $5.450$ & $5.650$ & $4.681$ \\
\textbf{HIR}                     &    $8.300$     & $4.450$   &    $3.900$     & $1.300$ &  $2.200$  & $3.150$ & $6.200$ & $7.800$ & $4.663$      \\
\textbf{DPO}                     &     \bm{$9.600$}    & $7.000$   &    $5.100$     &      $2.300$  &  $1.600$  & $\bm{5.600}$    &   \bm{$9.100$}    & $8.900$ & $6.150$  \\
\textbf{LORRA}                      &     $6.800$    & $5.700$   &    $2.300$     &      $1.800$  &  $2.300$  & $4.050$    &   $5.150$    & $6.700$ & $4.350$  \\
\textbf{LORRA-Pref}               &     $8.800$    & $6.950$   &    $5.100$     &      $1.400$  &  $2.100$  & $3.800$    &   $7.450$    & $8.000$ & $5.450$   \\
\rowcolor{gray!30}
\textbf{RAHF-Dual}                 &     $9.500$&$7.630$&\bm{$5.200$}&\bm{$3.400$}&$2.600$&$4.030$&$8.300$&$8.900$ & \bm{$6.195$}  \\
\rowcolor{gray!30}
\textbf{RAHF-SCIT}                     &     $9.150$    & \bm{$8.000$}   &    $3.600$     &      $2.400$  &  $2.200$  & $4.000$    &   $8.700$    & \bm{$9.350$} & $5.925$   \\
\hline
\multicolumn{1}{l}{\textbf{Turn-2}} \\
\hline
\textbf{Preferred-SFT}               &     $4.900$    & $7.000$   &    $2.700$     &      $1.100$  &  $1.900$  & $2.900$    &   $6.400$    & $8.200$ & $4.388$    \\
\textbf{RLHF-PPO}            &     $5.500$    & $7.500$   &    \bm{$4.700$}     &      $1.500$  &  \bm{$2.600$}  & $4.300$    &   $6.600$    & $6.400$ & $4.888$  \\
\textbf{HIR}                      & $     6.500$ & $      5.750$ & $      1.869$ & \bm{$      1.900$} & $      2.550$ & $      2.500$ & $      5.650$ & $      8.650$ & $4.421$     \\
\textbf{DPO}                      & \bm{$6.700$} & $      7.600$ & $      2.700$ & $      1.400$ & $      2.300$ & $      3.300$ & $      8.250$ & \bm{$      9.400$} & $5.206$     \\
\textbf{LORRA}                      & $     6.050$ & $      6.550$ & $      2.000$ & $      1.200$ & $      2.400$ & $      4.550$ & $      6.300$ & $      6.650$ & $4.463$  \\
\textbf{LORRA-Pref}               & $     5.800$ & $      7.100$ & $      3.200$ & $      1.400$ & $      1.500$ & \bm{$5.500$} & $      6.800$ & $      8.600$ & $4.988$  \\
\rowcolor{gray!30}
\textbf{RAHF-Dual}                & $6.650$&\bm{$7.850$}&$4.200$&$1.400$&$2.300$&$3.500$&$7.800$&$8.510$ &\bm{$5.276$}  \\
\rowcolor{gray!30}
\textbf{RAHF-SCIT}                     & $     5.000$ & $      7.300$ & $      3.600$ & $      1.700$ & $      1.700$ & $      3.700$ & \bm{$      8.300$} & \bm{$      9.400$} & $5.088$  \\
\hline
\multicolumn{1}{l}{\textbf{Final}} \\
\hline
\textbf{Preferred-SFT}               & $     6.700$ & $      6.700$ & $      3.400$ & $      1.650$ & $      2.000$ & $      3.650$ & $      6.450$ & $      7.625$ & $4.772$    \\
\textbf{RLHF-PPO}            & $     6.625$ & $      6.800$ & $      4.250$ & $      1.700$ & \bm{$      2.700$} & $      4.150$ & $      6.025$ & $      6.025$ & $4.784$ \\
\textbf{HIR}                      & $     7.400$ & $      5.100$ & $      2.885$ & $      1.600$ & $      2.375$ & $      2.825$ & $      5.925$ & $      8.225$ & $4.541$     \\
\textbf{DPO}                     & \bm{$     8.150$} & $      7.300$ & $      3.900$ & $      1.850$ & $      1.950$ & $      4.450$ & \bm{$      8.675$} & $      9.150$ & $5.678$   \\
\textbf{LORRA}                     & $     6.425$ & $      6.125$ & $      2.150$ & $      1.500$ & $      2.350$ & $      4.300$ & $      5.725$ & $      6.675$ & $4.407$  \\
\textbf{LORRA-Pref}               & $     7.300$ & $      7.025$ & $      4.150$ & $      1.400$ & $      1.800$ & \bm{$      4.650$} & $      7.125$ & $      8.300$ & $5.219$   \\
\rowcolor{gray!30}
\textbf{RAHF-Dual}                & $8.075$&\bm{$7.740$}&\bm{$4.700$}&\bm{$2.400$}&$2.450$&$3.765$&$8.050$&$8.705$ &\bm{$5.736$}  \\
\rowcolor{gray!30}
\textbf{RAHF-SCIT}                     & $     7.075$ & $      7.650$ & $      3.600$ & $      2.050$ & $      1.950$ & $      3.850$ & $      8.500$ & \bm{$      9.375$} & $5.506$  \\
\Xhline{1pt} % 设置粗细为2pt
\end{tabular}}
\caption{Results of MT-Bench.
}
\label{tab:mt-details}   
\end{table*}

\section{Qualitative Examples}
\begin{figure*}[t]
    \centering
    \includegraphics[width=0.99\linewidth]{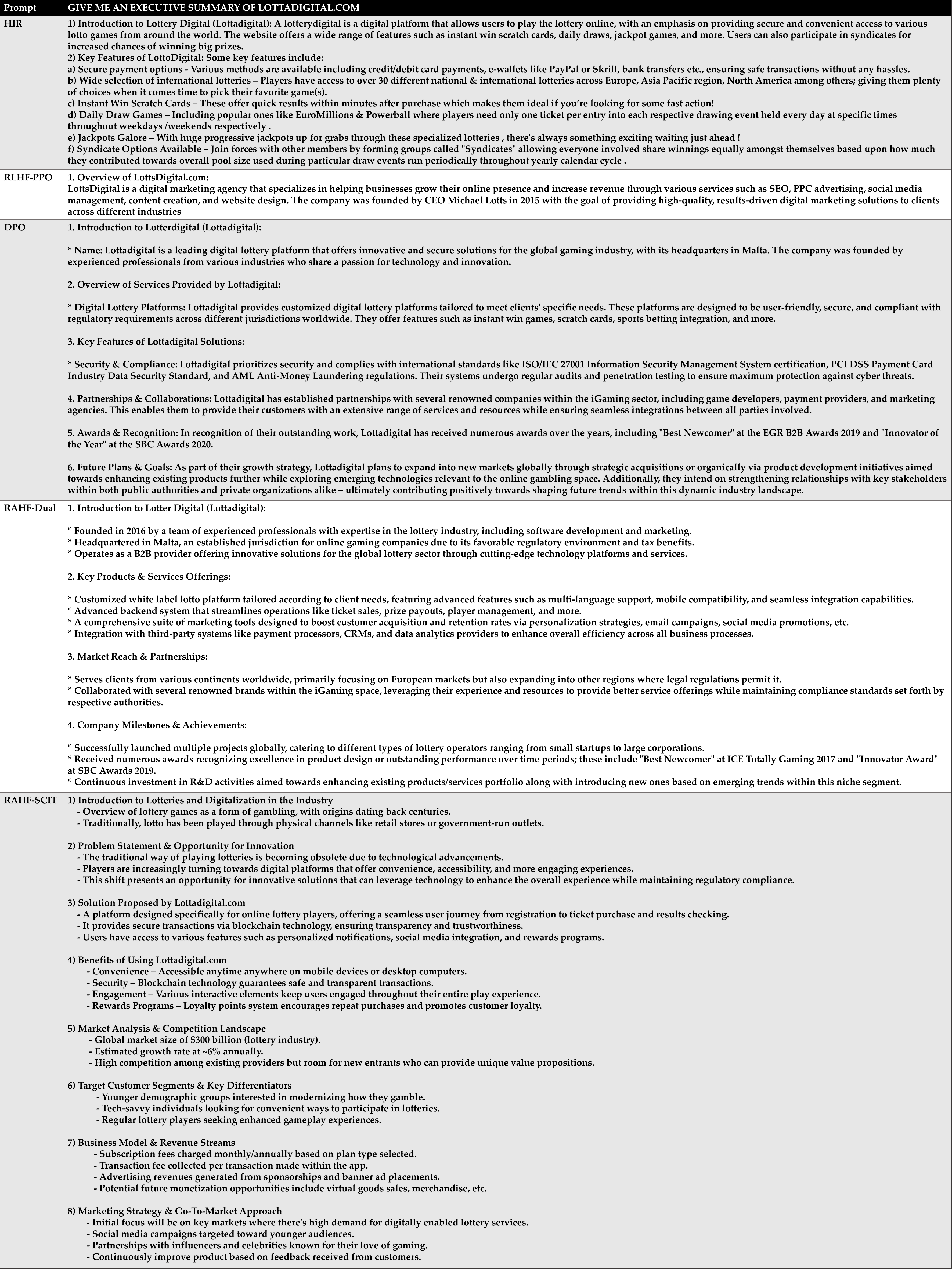}
    \caption{RAHF-Dual and RAHF-SCIT are more comprehensive and insightful compared to HIR, RLHF-PPO, and DPO. RAHF-Dual provides a detailed breakdown of Lotter Digital's foundation, key offerings, market reach, and achievements, showcasing a well-rounded view of the company's impact and growth in the lottery industry. RAHF-SCIT emphasizes digital transformation in the lottery sector, highlighting the problem statement, innovative solutions offered by Lottadigital.com, and the benefits and market potential of these solutions. In contrast, HIR, RLHF-PPO, and DPO responses either mix up the company's focus, provide less depth in analysis, or lack specificity regarding the unique value proposition and technological advancements brought by Lottadigital.com.}
    \label{Appendix: case1}
\end{figure*}

\begin{figure*}[t]
    \centering
    \includegraphics[width=0.99\linewidth]{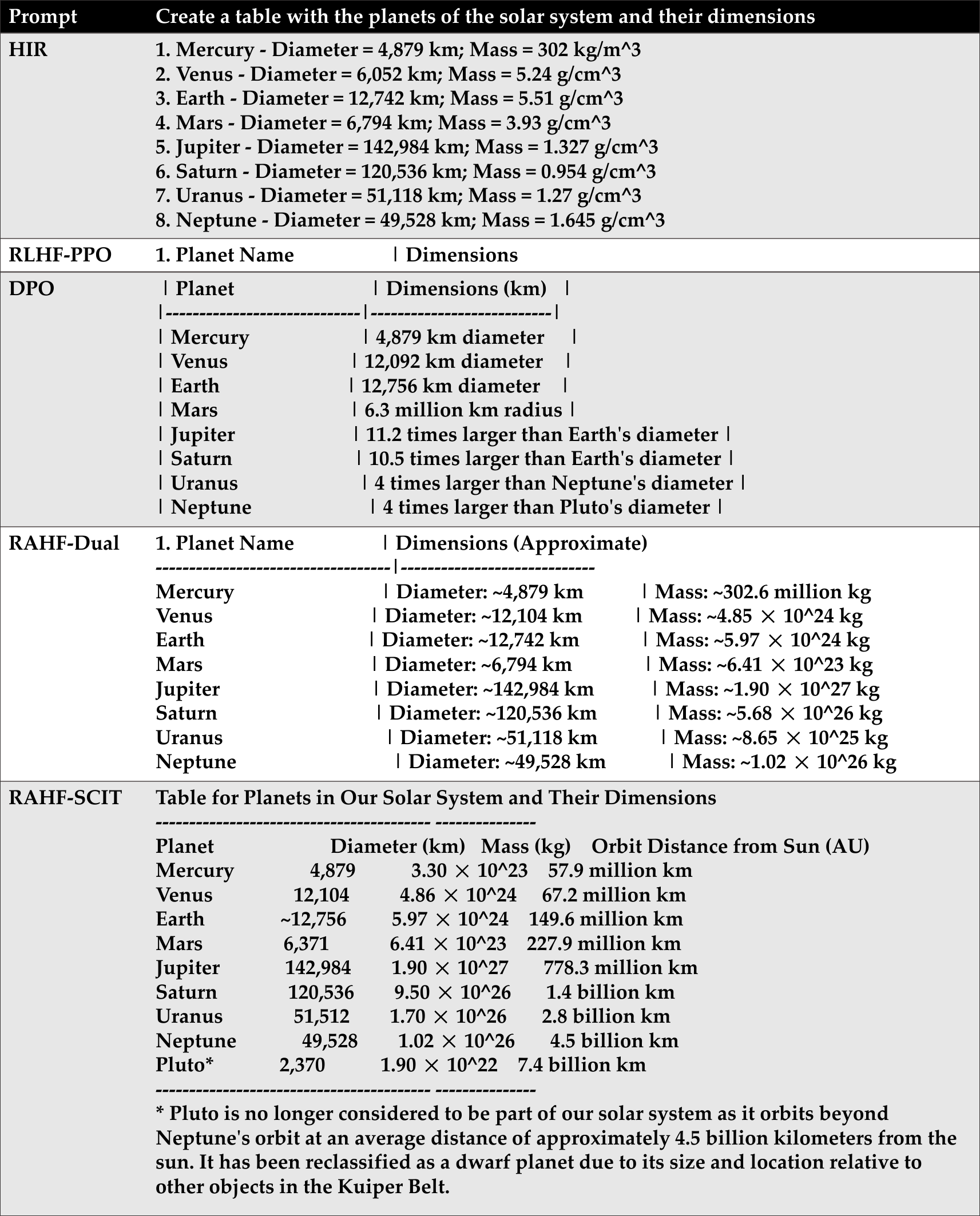}
    \caption{RAHF-Dual and RAHF-SCIT provide comprehensive, structured data with clear, consistent formatting, and include additional relevant details such as mass and orbit distance from the Sun. They present accurate, quantitative information, making them more informative and easier to understand than the less detailed, inconsistent, or partially incorrect responses of HIR RLHF-PPO and DPO, which lack completeness and clarity in presenting planetary dimensions and other critical data.}
    \label{Appendix: case2}
\end{figure*}

Figure \ref{Appendix: case1} and Figure \ref{Appendix: case2} present qualitative examples of RAHF compared with baselines in dialogue tasks.

\clearpage


\label{appendix:F}
\end{document}